\documentclass[10pt,twocolumn,letterpaper]{article}

\usepackage[pagenumbers]{cvpr} 

\usepackage[dvipsnames]{xcolor}

\definecolor{cvprblue}{rgb}{0.21,0.49,0.74}
\usepackage[pagebackref,breaklinks,colorlinks,citecolor=cvprblue]{hyperref}
\usepackage{cite}
\usepackage{amsmath,amssymb,amsfonts}
\usepackage{algorithmic}
\usepackage{graphicx}
\usepackage{textcomp}
\usepackage{xcolor}
\usepackage{algorithm}
\usepackage{algorithmic}
\usepackage{caption}
\usepackage{multirow}
\usepackage{array}
\usepackage{url}

\usepackage{etoolbox}
\makeatletter
\patchcmd{\@makecaption}{\scshape}{}{}{}
\patchcmd{\@makecaption}{\\}{.\ }{}{}
\makeatother

\def\BibTeX{{\rm B\kern-.05em{\sc i\kern-.025em b}\kern-.08em
		T\kern-.1667em\lower.7ex\hbox{E}\kern-.125emX}}
\usepackage{caption}
\def\paperID{10043} 
\def\confName{CVPR}
\def\confYear{2024}

\title{Boost Adversarial Transferability by Uniform Scale and Mix Mask Method}

\author{
Tao Wang, Zijian Ying, Qianmu Li, zhichao Lian\\
Nanjing University of Science and Technology\\
{\tt\small 122106010829@njust.edu.cn, zjying@njust.edu.cn, qianmu@njust.edu.cn, lzcts@163.com}
}

\begin{document}
\maketitle
\begin{abstract}
Adversarial examples generated from surrogate models often possess the ability to deceive other black-box models, a property known as transferability. 
Recent research has focused on enhancing adversarial transferability, with input transformation being one of the most effective approaches. 
However, existing input transformation methods suffer from two issues. 
Firstly, certain methods, such as the Scale-Invariant Method, employ exponentially decreasing scale invariant parameters that decrease the adaptability in generating effective adversarial examples across multiple scales. 
Secondly, most mixup methods only linearly combine candidate images with the source image, leading to reduced features blending effectiveness. 
To address these challenges, we propose a framework called Uniform Scale and Mix Mask Method (US-MM) for adversarial example generation. 
The Uniform Scale approach explores the upper and lower boundaries of perturbation with a linear factor, minimizing the negative impact of scale copies. 
The Mix Mask method introduces masks into the mixing process in a nonlinear manner, significantly improving the effectiveness of mixing strategies. 
Ablation experiments are conducted to validate the effectiveness of each component in US-MM and explore the effect of hyper-parameters.
Empirical evaluations on standard ImageNet datasets demonstrate that US-MM achieves an average of $7\%$ better transfer attack success rate compared to state-of-the-art methods. 
\end{abstract}    
\section{Introduction}
\label{sec:intro} \indent

Deep neural networks (DNNs) have demonstrated remarkable success in computer vision tasks. However, recent research has revealed that adding imperceptible perturbations to images can deceive these models\cite{goodfellow2014explaining}, posing serious security concerns. For example, such attacks can compromise personal safety in autonomous driving\cite{pouyanfar2018survey} or cause issues with facial recognition systems\cite{sharif2016accessorize}. One concerning phenomenon is that adversarial examples crafted to deceive one model can also fool other models, even if they have different architectures and parameters. This transferability of adversarial examples has garnered widespread attention.
Attackers can exploit this property without having detailed knowledge of the target model, making it a black-box attack. While these adversarial example generation methods were initially proposed for white-box attacks where the attacker has complete knowledge of the target model, they can still suffer from overfitting issues with the source model. This can result in weak transferability when these adversarial examples are used to attack other black-box models.

\begin{figure}[tbp]
	\centerline{\includegraphics[width=1.0\columnwidth]{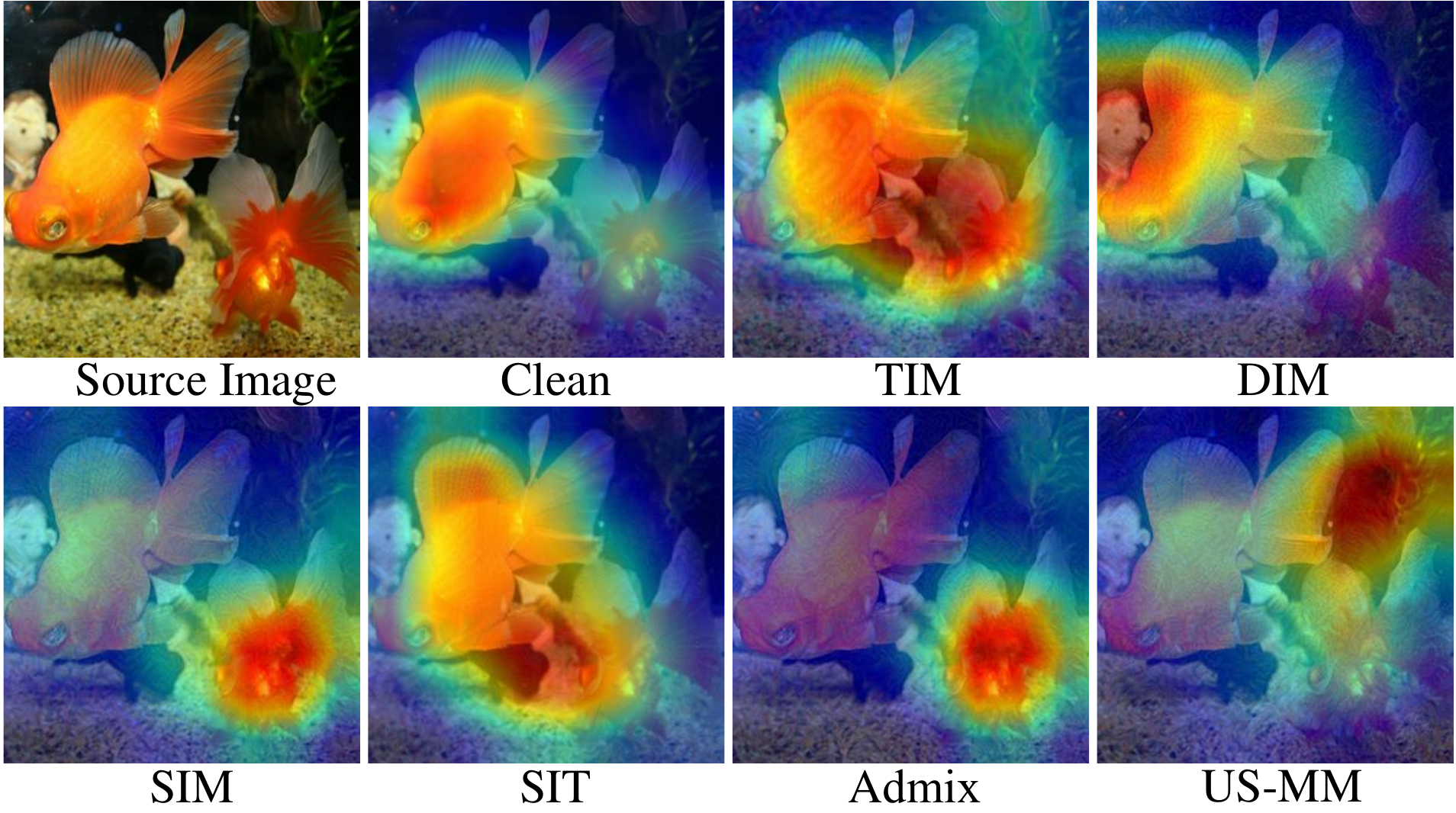}}
	\caption{A collection of heatmaps about the source image and adversarial examples crated by DIM \cite{xie2019improving}, TIM\cite{dong2019evading}, SIM\cite{lin2019nesterov}, SIT\cite{wang2023structure}, Admix\cite{wang2021admix} and our proposed US-MM. The red regions are of significance in model prediction.}
	\label{Fg_1_heatmap}
\end{figure}

For boosting the transferability of adversarial examples in black-box setting, various approaches have been proposed.
The optimizer-based methods\cite{dong2018boosting,lin2019nesterov,wang2021enhancing,gao2020patch,li2020towards,qin2022boosting} improve adversarial transferability by optimizing the research routine and the target points.
The mid-layer-based methods\cite{zhou2018transferable,huang2019enhancing,ganeshan2019fda,wu2020boosting,zhang2022improving} try to decrease the influence from the specific model.
These methods design the loss function elaborately to enlarge the distance of features in intermediate layers between adversarial example to improve the generalization.
The ensemble-model methods\cite{liu2016delving,dong2018boosting,li2020learning,xiong2022stochastic} utilize the generalization from multiple networks.
These models attack multiple networks at the same time and maximize the sum of model losses. 
The methods based on input transformation try to introduce transformed images, e.g. scale transformation, to craft adversarial examples.
Current works indicate that adding additional information from those transformed images can enhance the adversarial transferability\cite{long2022frequency,zhang2023improving}.
These methods achieved leadership in computational performance and attack effectiveness.

However, in the current state-of-the-art transformation-based attacks, there is a lack of focus on the importance of transformation factors. These factors play a crucial role in the generation of adversarial examples and greatly impact the effectiveness of the corresponding methods. For instance, the Scale-Invariant Method (SIM)\cite{lin2019nesterov}, one of the top-performing methods, uses an exponential scale to incorporate features from different scale-levels into the target image. While this strategy has been shown to enhance the transferability of adversarial examples, the use of multiple scales can limit their effectiveness when the number of scales increases. As a result, SIM requires careful selection of scale-invariant factors to achieve optimal performance.
The most recent model, Admix\cite{wang2021admix}, takes advantage of the mixup strategy, where information from images of other classes is introduced to further improve transferability. However, this mixup strategy is limited by its linear approach, which weakly adds information without adaptation. Additionally, this linear mix can also damage some pixel regions in the source image, leading to limited transferability of the adversarial examples.

To address these issues, we propose a novel and flexible attack method called the Uniform Scale and Mix Mask Method (US-MM). The US component overcomes the limitations of SIM by uniformly sampling scale values within an interval. The MM component improves upon the mixup strategy by using mix masks for multiplication instead of addition. Each component can directly improve the transferability of adversarial examples, and the combination of the two can further enhance performance. Additionally, the two parts of US-MM can be integrated into other attack methods separately to achieve even stronger results.

Our contributions are summarized as follows:

\begin{itemize}
	\item We propose a novel method called Uniform Scale Method (US) which uniformly samples scale values within an interval while considering the upper and lower bounds for the perturbations. This approach can effectively address the issue from large number of scale copies.
	\item We propose a novel non-linear mixup strategy, namely Mix Mask Method(MM), that incorporates the mix image into the mask. This approach can effectively enhance information addition and overcome the issue of damaged regions in the source image.
	\item We conduct ablation experiments to validate the effectiveness of both the US and MM methods. The results showed that both methods can significantly improve the transferability of adversarial examples individually. Additionally, we explored the effect of hyper-parameters on the performance of these methods.
	\item We conduct a comparison experiment on the benchmark dataset ImageNet, using 5 state-of-the-art baselines. Experimental results show that the proposed US-MM method achieves significantly higher transfer attack success rates compared to the baselines. Under optimal settings, US-MM achieves an average improvement of 7.3\% over the best-performing baseline method.
\end{itemize}

\section{Related Work}  \indent
\label{sec:related}

Attack methods can be categorized into two types based on the attacker's knowledge of the victim model: white-box attacks \cite{madry2017towards} and black-box attacks. In white-box attacks, the attacker has access to all information about the target model, such as its architecture and parameters, to generate adversarial examples. In contrast, black-box attacks only allow attackers to have query permissions.

\subsection{White-box Adversarial Attack} \indent

Fast Gradient Sign Method (FGSM)\cite{goodfellow2014explaining} sets the optimization objective to maximize the loss of classification function and makes the sign of input gradient as noise for benign image. Basic Iterative Method (BIM)\cite{kurakin2016adversarial} extends the idea of FGSM by applying multiple iterations of small perturbations to achieve better white-box attack performance. DeepFool\cite{moosavi2016deepfool} leads adversarial examples close to decision boundary continuously in iterations. Carlini and Wagner attacks (C\&W)\cite{carlini2017towards} is an optimizer-based method which aims to minimize the distance between adversarial example and benign image subject to classification error. 

Although these methods can achieve nearly 100\% success rates in white-box attack setting, when tested on other black-box models, the adversarial examples show weak transferability due to overfitting with the source model.

\subsection{Black-box Adversarial Attack} \indent

It is more challenging in black-box attack scenarios because attackers are absolutely ignorant of victim model but only obtain model output. There are two sorts of black-box attack algorithms. One is query-based attacks\cite{su2019one,bhagoji2018practical,ilyas2018black,guo2019simple,shi2020polishing} while the other is transfer-based attacks. Query-based attacks design query samples purposefully and optimize adversarial noise based on query results. However, it is impractical in physical world because of the huge amount of query operations. 

Instead, based on the phenomenon that adversarial examples generated for one model might mislead another model, transfer-based attacks works by attacking a local surrogate model. To enhance transferability, existing transfer-based attacks usually utilize several avenues to craft adversarial examples. 

\textbf{Optimizer-based attacks.} Dong \textit{et al.}\cite{dong2018boosting} use momentum to help escape from poor local minima in multiple iterations, denoted as MI-FGSM. Lin \textit{et al.}\cite{lin2019nesterov} substitute the image which moves forward in the direction of momentum for source image to calculate gradient with the main idea of looking ahead. Wang \textit{et al.}\cite{wang2021boosting} estimate momentum on several samples which crafted on previous gradient’s direction repeatedly. Wang \textit{et al.}\cite{wang2021enhancing} utilize the average value of gradient difference between source image and surrounding samples to swap adversarial examples, achieving higher transferability.

\textbf{Mid-layer-based attacks.} 
Zhou \textit{et al.}\cite{zhou2018transferable} introduce two terms into loss function, where the first term is used to maximize the distance of feature maps between the input image and adversarial examples while the second term aim to reduce high-frequency disturbances. 
Huang \textit{et al.}\cite{huang2019enhancing} shift the adversarial noise to enlarge the distance of specific layer in DNNs between the benign image and adversarial examples. 
Ganeshan \textit{et al.}\cite{ganeshan2019fda} design a novel loss function which reduces the activation of supporting current class prediction and enhances the activation of assisting other class prediction.
Wu \textit{et al.}\cite{wu2020boosting} optimize adversarial examples by maximizing the distance of attention map between the adversarial examples and the original image.

\textbf{Ensemble-model attacks.} Liu \textit{et al.}\cite{liu2016delving} argue that the adversarial examples might have the stronger transferability if they can the cheat more networks. They attack multiple models simultaneously and aim to maximize the sum of model losses. 
Li \textit{et al.}\cite{li2020learning} introduce dropout layers into source model and acquire ghost networks by setting different parameters. In each iteration, the surrogate model is selected randomly from network collection, known as longitudinal ensemble. 
Xiong \textit{et al.}\cite{xiong2022stochastic} tune the ensemble gradient in order to reduce the variance of ensemble gradient for each single gradient.

\textbf{Input transformation based attacks.} Xie \textit{et al.}\cite{xie2019improving} propose the first attack method based on input transformation. They resize the input image randomly and expand it to a fixed size by filling pixels. Dong \textit{et al.}\cite{dong2018boosting} use a set of translated images to optimize adversarial examples. To reduce computation complexity, they apply convolution kernel to convolve the gradient. Lin \textit{et al.}\cite{lin2019nesterov} assume the scale-invariant property of DNNs and propose an attack method working by calculating the gradient by several scaled copies, denoted as SIM. Wang \textit{et al.}\cite{wang2021admix} observe that introducing the information of images in other categories during generating examples can improve the transferability significantly. They propose an Admix method which mixes source image and the images with different labels. Wang \textit{et al.} \cite{wang2023structure} divide input image into several regions and apply various transformations onto the image blocks while retaining the structure of image.

\section{Preliminary} \indent

In this section, we will first define the notations used for generating adversarial examples. Then, we will discuss the limitations of the SIM and Admix methods and state the motivation behind our work.

\subsection{Problem Settings} \indent

Adversarial attack tries to create an adversarial example $x^{adv}$ from a benign image $x$. The victim model, typically a deep learning model, is denoted as $f$, with parameters $\theta$. The output of the victim model is $f(x,\theta) \in R^K$, where $K$ is the number of classes. The true label of $x$ is represented as $y$, and the loss function of $f$ is denoted as $J(f(x;\theta),y)$. The objective of the attack method is to generate an adversarial example within a specified constraint, such that the victim model misclassifies it. This can be achieved by generating $x^{adv}$, which satisfies $\Vert x^{adv}-x \Vert_p<\epsilon$, and results in $f(x^{adv};\theta)\neq f(x;\theta)$. The most common constraint used in adversarial attacks is the $L_{\infty}$ norm.

\subsection{Motivation} \indent

To enhance adversarial transferability, the Scale-Invariant Method (SIM) \cite{lin2019nesterov} utilizes the gradient from multiple scaled images to generate adversarial examples. The key concept of SIM is the scale-invariant property of Deep Neural Networks (DNNs), which means that the network produces similar predictions for inputs of different scales. The core of SIM is using scaling transformation $S(x)$ to modify the input image, which is represented as follows:
\begin{equation}
	S_i(x)= x/2^i,  
\end{equation}
where $i$ is the indicator of scale copies. Description for the updating strategy of SIM is
\begin{equation}
	x^{adv}_{t+1}=x^{adv}_t+\alpha*sign(\frac{1}{m}\sum_{i=0}^{m-1}\nabla_{x^{adv}_t}J(f(S_i(x^{adv}_t);\theta),y)), 
\end{equation}
where $m$ is the number of scale copies, $\nabla(\cdot)$ is the gradient, $sign(\cdot)$ is the direction function, $\alpha$ is updating rate and $t$ is the iteration round indicator.

\begin{figure}[htbp]
	\centerline{\includegraphics[width=1.0\columnwidth]{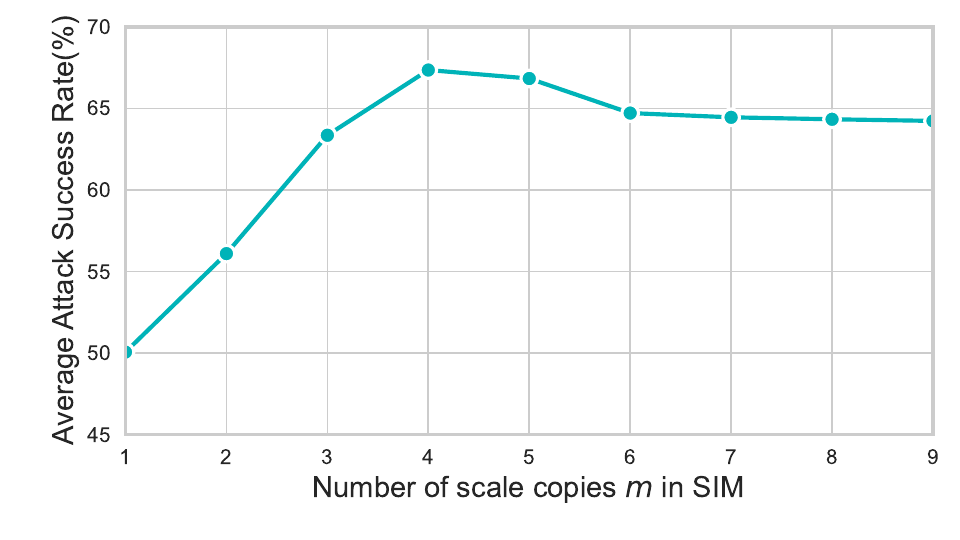}}
	\caption{Average attack success rates (\%) of SIM when attacking five pretrained models. The examples are crafted on Inc-v3 model.}
	\label{fig1}
\end{figure}

SIM is based on the assumption of the scale-invariant property of DNNs, where the model has similar losses for a certain degree of scale-changed images as it does for the original image. SIM uses the scaling transformation $S_i(x)= x/2^i$ and a hyper-parameter $m$ to limit the number of scale copies, with a higher $m$ representing more general feature information. However, when $m$ is greater than a certain degree, SIM's performance starts to decline, as shown in Figure \ref{fig1}. This is because the pixel values in $S_i(x)$ tend to 0 when $i$ is a large integer, resulting in a nearly black image. This can negatively affect the generalization of features and lead to a decrease in performance. Therefore, we believe that using the gradient of such scaled images to generate adversarial noise can reduce transferability, and a lower bound for the scale change is necessary.

Unlike SIM, the mixup strategy enhances adversarial transferability by introducing features from other classes. Admix \cite{wang2021admix}, the current state-of-the-art mixup strategy, addresses the issue of the mixup portion of the mixed image using linear weights $\eta$. This is represented as $x_{mixed}=x+\eta\cdot x^{\prime}$, where $x$ is the input image and $x^{\prime}$ is the mixed image. However, there are two problems arise.

The first problem with the Admix method is that for a random pixel $P_{x^{\prime}}$ in $x^{\prime}$, even if $\eta$ decreases its value, it is still unpredictable whether the pixel value will be greater than the corresponding pixel $P_{x}$ in $x$. This means that the condition $P_{x}<\eta\cdot P_{x^{\prime}}$ will always be true, unless $\eta$ is very close to 0. However, mixed image will lose efficacy when $\eta$ is too small. As a result, the mixed image $x_{mixed}$ will have a larger portion from the image of another category in some pixel positions, which can significantly disrupt the feature information of the original image $x$.

The second issue with Admix is that the pixel values in $x^{\prime}$ are always positive, which only increases input diversity in the positive direction. This results in a limited range of mixing options. It may be more effective to mix the image in a negative direction as well.

\section{Methodology}
\label{sec:methodology} \indent

In this section, we first introduce the Admix method. Then, we present our proposed methods, the Uniform Scale Method and Mix Mask Method, and provide an algorithm for a better understanding of the proposed approach.

\begin{figure*}
\centerline{\includegraphics[width=2.0\columnwidth]{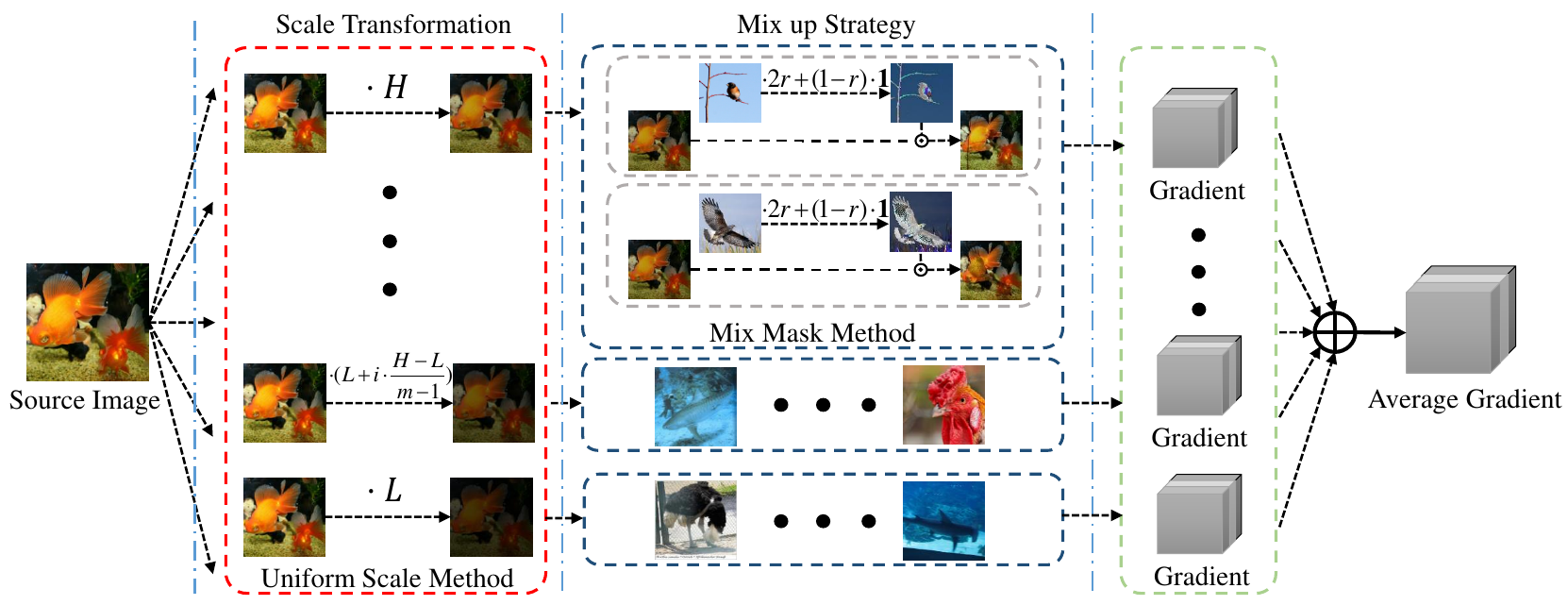}}
\caption{Illustration of Uniform Scale and Mix Mask Method (US-MM). Firstly, the source image is scaled uniformly, generating multiple scale copies. Then, mix masks are crated on sampled mix images and applied on each scale copy. Finally, the gradient is calculated by all transformed images. }
\label{Fg_4_US_MM_graph}
\end{figure*}

\subsection{Admix} \indent

Admix \cite{wang2021admix} uses a set of admixed images to calculate the average gradient. It first randomly choose several images from other categories. Then, for every sampled image $x^{\prime}$, Admix gets the mixed image $\tilde{x}$ by adopting a linear mix method on original image $x$ as follows:
\begin{equation}
	\tilde{x}=\gamma \cdot x+\eta^{\prime}\cdot x^{\prime}=\gamma \cdot(x+\eta \cdot x^{\prime}), 
\end{equation}
where $\gamma$ and $\eta^{\prime}$ are the mix portions of the original image and sampled image
in the admixed image respectively satisfying $0 \leq\eta^{\prime} < \gamma \leq 1$ and $\eta$ is computed by $\eta=\eta^{\prime} / \gamma$.

Admix also keeps the rule of SIM, uses $S(x)$ to obtain the value of $\gamma$.
Thus, Admix works as follows:
\begin{align}
	&\bar{g}_{t+1}=  \notag  \\ 
	&\frac{1}{m_1*m_2}\sum_{x^{\prime}\in X^{\prime}} \sum_{i=0}^{m_1-1} \nabla_{x^{adv}_t}J(f(S_i(x^{adv}_t+\eta\cdot x^{\prime});\theta),y), 
\end{align}
\begin{equation}
	x^{adv}_{t+1}=x^{adv}_t+\alpha*sign(\bar{g}_{t+1}),
\end{equation}
where $m_1$ is the number of admixed images for each $x^{\prime}$ and $X^{\prime}$ is an image collection containing $m_2$ randomly sampled images which have different labels with $x$.

\subsection{Uniform Scale Method} \indent

The most straightforward solution to address the issue with SIM \cite{lin2019nesterov} is to define a lower bound to avoid generating meaningless or nearly black images. This can be achieved by also considering an upper bound in the formulation of $S_i(x)$, which can be reformed as 
\begin{equation}
	S_i(x,L,H)=(L+\frac{H-L}{2^i})\cdot x, \label{eq000}
\end{equation}
where $x$ is the input image, $L$ is the lower bound for the scale, and $H$ is the upper bound for the scale. Both $L$ and $H$ are floating-point numbers between 0 and 1, with the condition that $L\leq H$. These parameters can be used to control scale range and enhance the adversarial transferability within a suitable scope. In the special case where $L=0$ and $H=1$, the equation is reduced to the original SIM method.

However, because of the exponential scale function, when number of scale copies is great, the majority of scale copies are close to $L \cdot x$, which have similar gradient information, denote as $g_L$. Finally, the calculated average gradient tends to $g_L$, decreasing input diversity instead.

To overcome this problem, we further utilize the uniform scale with a convert function $U_i(x,m,L,H)$ to generate scale copies, which we called Uniform Scale Method (USM). USM obtains scale values uniformly from the range between scale lower bound $L$ and upper bound $H$, which is
\begin{equation}
	U_i(x,m_{us},L,H)=(L+i*\frac{H-L}{m_{us}-1})\cdot x, \label{eq1}
\end{equation}
where $x$ is the input image and $m_{us}$ is a positive integer to present the number of uniform scale copies. Particularly, we stipulate that $U_i(x,m_{us},L,H)=H \cdot x$ when $m_{us}=1$.

\subsection{Mix Mask Method} \indent

To address the limitations of linear mixup, we propose the Mix Mask Method (MM), which works by generating a mix mask from an image of a different category and applying it to the input image. This method has two main improvements: first, the transformation range is related to the per-pixel value of the source image, and second, the transformation contains both positive and negative directions. In the first step of MM, a mix mask is generated according to the mix image using the following equation:
\begin{equation}
	M_{mix}=(1-r) \cdot \boldsymbol{1}+2r \cdot x^{\prime}, \label{eq2}
\end{equation}
where $M_{mix}$ is the mask, $r$ is the mix range size, $x^{\prime}$ is the mix image and $\boldsymbol{1}$ is an all one matrix same shape with $x^{\prime}$. Because the images are normalized to [0, 1], the value of each element in $M_{mix}$ is mapped into [1-r, 1+r].

Then, mask $M_{mix}$, which contains the information of the mix image, can be utilized to influence the source image $x$, which is
\begin{equation}
	x^{m}=M_{mix} \odot x, \label{eq3}
\end{equation}
where $x^{m}$ is the mixed image generated by the source image and the mask, and $\odot$ is element-wise product. 

In MM, the transformation range of per pixel in source image is limited within a symmetric interval by applying mix mask, which means a kind of reliable and bidirectional transformation measure. Then MM can introduce features from other categories of images more effectively than linear ways.

\subsection{Algorithm of US-MM Method} \indent

US-MM method contains two parts, scale transformation and mix up strategy, same as Admix. The structure of our US-MM method is exhibited in Figure \ref{Fg_4_US_MM_graph}.   The pseudo-code of the process of Uniform Scale and Mix Mask Method is summarized in Algorithm \ref{alg1}. Note that our US component can replace SIM in any appropriate situation and it is easy to integrate our MM component into other transfer-based attacks.

\begin{algorithm}
	\caption{ Uniform Scale and Mix Mask Method}
	\label{alg1}
	\begin{algorithmic}[1]
		\REQUIRE A classifier $f$ with parameter $\theta$ and loss function $J$
		\REQUIRE A benign image $x$ with ground-truth label $y$
		\REQUIRE The maximum perturbation $\epsilon$ and number of iterations $T$.
		\REQUIRE Number of uniform scale copies $m_{us}$, scale lower bound $L$, scale upper bound $H$, 
		\REQUIRE Number of mix images $m_{mix}$, mix range size $r$
		\ENSURE An adversarial example $x^{adv}$
		\STATE $ \alpha = \frac{\epsilon}{T}; x^{adv}_0=x$
		\FOR{$t=0$ to $T-1$ }
		\STATE $G=0$
		\FOR{$i=0$ to $m_{us}-1$ }
		\STATE $x^{scaled}_i=U_i(x^{adv}_t,m_{us},L,H)$
		\FOR{$j=0$ to $m_{mix}-1$ }
		\STATE Get a mix mask $M_{mix}$ by Eq.\eqref{eq2}
		\STATE $x^{m}_{i,j}=M_{mix}\odot x^{scaled}_i$
		\STATE $x^{m}_{i,j}=Clip(x^{m}_{i,j},0,1)$
		\STATE $G=G+\nabla_{x^{m}_{i,j}}J(f(x^{m}_{i,j};\theta),y)$
		\ENDFOR
		\ENDFOR
		\STATE $x^{adv}_{t+1}=x^{adv}_t+\alpha*sign(G)$
		\ENDFOR
		\RETURN $x^{adv}=x^{adv}_T$
	\end{algorithmic}
\end{algorithm}

\section{Experiments} \indent

In this secion, we conduct experiments to verify the effectiveness of our proposed approach. We first specify the setup of the experiments. Then, we do ablation study to explore the role of different scale lower bound $L$, scale upper bound $H$ and mix range size $r$. We also display the effectiveness of our two methods. Finally, we report the results about attacking several pretrained models with baseline methods and US-MM.
\subsection{Experiment Setup}
\textbf{Dataset}. We evaluate all methods on 1000 images from ILSVRC 2012 validation set\cite{russakovsky2015imagenet} provided by Lin et al.\cite{lin2019nesterov}.

\noindent \textbf{Models}. We study six pretrained models based on ImageNet, i.e. Inception-v3 (Inc-v3)\cite{szegedy2016rethinking}, VGG16\cite{simonyan2014very}, ResNet50 (Res50)\cite{he2016deep} , DenseNet121 (Dense121)\cite{huang2017densely}, Inception-v4 (Inc-v4) and Inception-ResNet-v2 (IncRes-v2)\cite{szegedy2017inception}. All these models can be found in \footnote{\url{https://github.com/Cadene/pretrained-models.pytorch}}.

\noindent \textbf{Baselines}. We choose five transformation-based attack methods as the baselines, i.e.  DIM\cite{xie2019improving}, TIM\cite{dong2019evading}, SIM\cite{lin2019nesterov}, SIT\cite{wang2023structure} and Admix\cite{wang2021admix}. SIT is the latest method. All attacks are integrated into MI-FGSM\cite{dong2018boosting}, which is the most classic method to improve transferability.

\noindent \textbf{Attack setting}. We follow the most settings in \cite{wang2021admix}. We set the maximum perturbation $\epsilon$ to 16 and number of iteration $T$ to 10. For MI-FGSM, we make momentum delay factor $\mu=1.0$. We set the probability of input transformation $p=0.7$ in DIM and use Gaussian kernel with size of $7 \times 7$ in TIM. Based on our study about SIM, we set scale copies $m=5$ to achieve the best performance. For Admix, except for the same setting as SIM, we set the number of mix images to 3 and mix ratio $\eta=0.2$. To keep the same computational complexity with Admix, we set the splitting number $s = 3$ and change the number of transformed images for gradient calculation $N$ to 15 in SIT.

\subsection{Ablation Studies} \indent

In this section, we study the attack performance with different value of hyper-parameters and verify the effectiveness of our two methods. To deducing the impact of randomness about MM, we set $r=0$ to degenerate US-MM into USM when we conduct experiments about scale lower bound $L$ and scale upper bound $H$. All  experiments only treat Inc-v3 as the victim model and attack methods are realized based on MI-FGSM.

\begin{table*}[t]
	\centering
	\small
	\begin{tabular}[c]{|c|c|cccccc|}
		\hline
		Model & Attack & Inc-v3& VGG16 & Res50 & Dense121 & Inc-v4 & IncRes-v2 \\
		\hline
		\hline
		\multirow{6}*{Inc-v3} & DIM & 94.5$^*$ & 48.4 & 40.3 & 42.7 & 39.1 & 34.4 \\
		~ & TIM & 99.8$^*$ & 56.2 & 43.4 & 48.9 & 36.4 & 30.1 \\
		~ & SIM & \bf{100.0}$^*$ & 69.3 & 64.6 & 68.5 & 67.3 & 64.5 \\
		~ & SIT & \bf{100.0}$^*$ & 88.1 & 81.8 & 80.8 & 82.3 & 74.8 \\
		~ & Admix & 99.9$^*$ & 80.1 & 75.3 & 81.3 & 81.6 & 81.9 \\
		~ & US-MM & \bf{100.0}$^*$ & \bf{91.9} & \bf{89.8} & \bf{91.5} & \bf{92.8} & \bf{93.0} \\ \hline
		\multirow{6}*{VGG16} & DIM & 43.4 & 99.9$^*$ & 59.1 & 59.5 & 42.8 & 30.3 \\
		~ & TIM & 45.5 & 99.6$^*$ & 66.4 & 68.8 & 43.8 & 31.0 \\
		~ & SIM & 78.9 & \bf{100.0}$^*$ & 83.2 & 84.9 & 78.6 & 65.5 \\
		~ & SIT & 60.1 & \bf{100.0}$^*$ & 87.6 & 84.9 & 62.0 & 41.5 \\
		~ & Admix & 85.1 & \bf{100.0}$^*$ & 89.7 & 92.2 & 88.1 & 76.9 \\
		~ & US-MM & \bf{92.4} & \bf{100.0}$^*$ & \bf{94.4} & \bf{95.1} & \bf{92.4} & \bf{83.1} \\ \hline
		\multirow{6}*{Res50} & DIM & 45.5 & 71.1 & 99.0$^*$ & 71.9 & 41.2 & 37.7 \\
		~ & TIM & 48.6 & 79.0 & \bf{100.0}$^*$ & 79.9 & 42.8 & 37.3 \\
		~ & SIM & 82.0 & 89.5 & \bf{100.0}$^*$ & 94.3 & 76.7 & 73.9 \\
		~ & SIT & 76.3 & 96.7 & \bf{100.0}$^*$ & 98.4 & 71.7 & 62.1 \\
		~ & Admix & 91.0 & 94.6 & \bf{100.0}$^*$ & 97.6 & 87.4 & 85.3 \\
		~ & US-MM & \bf{96.5} & \bf{98.4} & \bf{100.0}$^*$ & \bf{99.5} & \bf{92.9} & \bf{91.4} \\ \hline
		\multirow{6}*{Dense121} & DIM & 50.2 & 73.8 & 74.5 & 99.6$^*$ & 46.7 & 41.0 \\
		~ & TIM & 50.9 & 81.8 & 77.3 & \bf{100.0}$^*$ & 48.3 & 40.8 \\
		~ & SIM & 82.1 & 91.2 & 92.9 & \bf{100.0}$^*$ & 79.0 & 75.5 \\
		~ & SIT & 81.3 & 97.9 & 99.2 & \bf{100.0}$^*$ & 78.5 & 68.2 \\
		~ & Admix & 90.3 & 95.8 & 96.4 & \bf{100.0}$^*$ & 89.4 & 86.6 \\
		~ & US-MM & \bf{96.2} & \bf{99.2} & \bf{99.4} & \bf{100.0}$^*$ & \bf{94.6} & \bf{92.9} \\ \hline
		\multirow{6}*{Inc-v4} & DIM & 45.5 & 56.5 & 39.3 & 44.1 & 90.5$^*$ & 37.3 \\
		~ & TIM & 42.9 & 59.1 & 44.5 & 51.3 & 96.6$^*$ & 34.6 \\
		~ & SIM & 79.0 & 78.4 & 68.2 & 76.6 & 99.7$^*$ & 76.2 \\
		~ & SIT & 83.2 & 90.8 & 78.7 & 80.3 & 99.8$^*$ & 70.4 \\
		~ & Admix & 88.1 & 84.9 & 78.5 & 84.8 & \bf{99.9}$^*$ & 85.8 \\
		~ & US-MM & \bf{95.8} & \bf{95.1} & \bf{90.2} & \bf{93.7} & 99.8$^*$ & \bf{93.4} \\ \hline
		\multirow{6}*{IncRes-v2} & DIM & 45.3 & 52.3 & 41.4 & 43.2 & 43.5 & 82.7$^*$ \\
		~ & TIM & 47.4 & 59.9 & 50.3 & 53.4 & 44.2 & 89.8$^*$ \\
		~ & SIM & 80.1 & 73.4 & 70.4 & 73.3 & 77.9 & \bf{99.9}$^*$ \\
		~ & SIT & 90.6 & 88.4 & 84.9 & 85.4 & 86.5 & 99.5$^*$ \\
		~ & Admix & 87.2 & 79.5 & 80.0 & 83.0 & 85.6 & 99.8$^*$ \\
		~ & US-MM & \bf{95.0} & \bf{90.1} & \bf{89.3} & \bf{90.9} & \bf{93.5} & 99.7$^*$ \\ \hline
	\end{tabular}
	\\ [0.2cm]
	\large
	\captionsetup{justification=justified}
	\caption{The attack success rates (\%) against six models by baseline attacks and our method. The best results are marked in bold and * represent it is white-box attack setting.}
	\label{big_table}
\end{table*}

\begin{figure}[tbp]
	\centerline{\includegraphics[width=1.0\columnwidth]{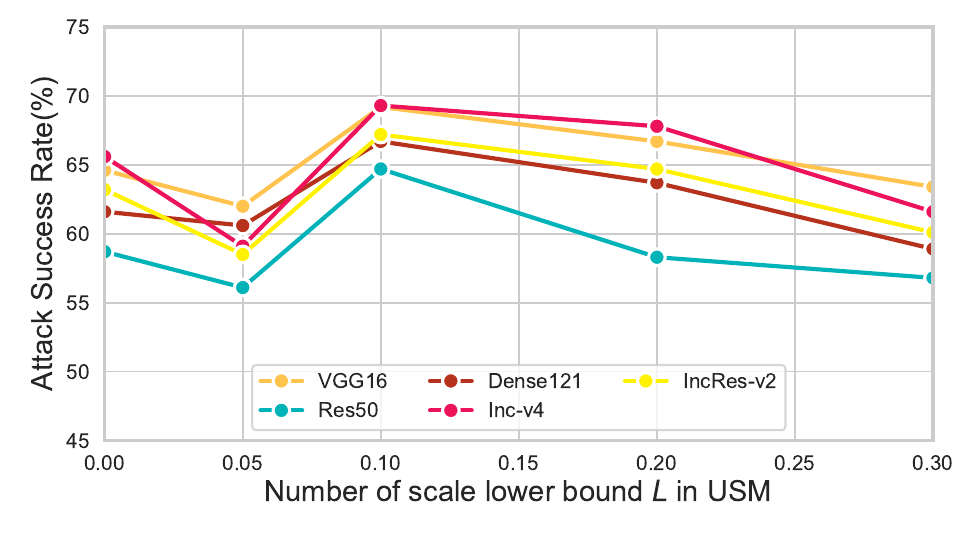}}
	\caption{Attack success rates (\%) of USM when attacking other five pretrained models for different scale lower bound $L$.}
	\label{Fg_5_lower}
\end{figure}

\begin{figure}[tbp]
	\centerline{\includegraphics[width=1.0\columnwidth]{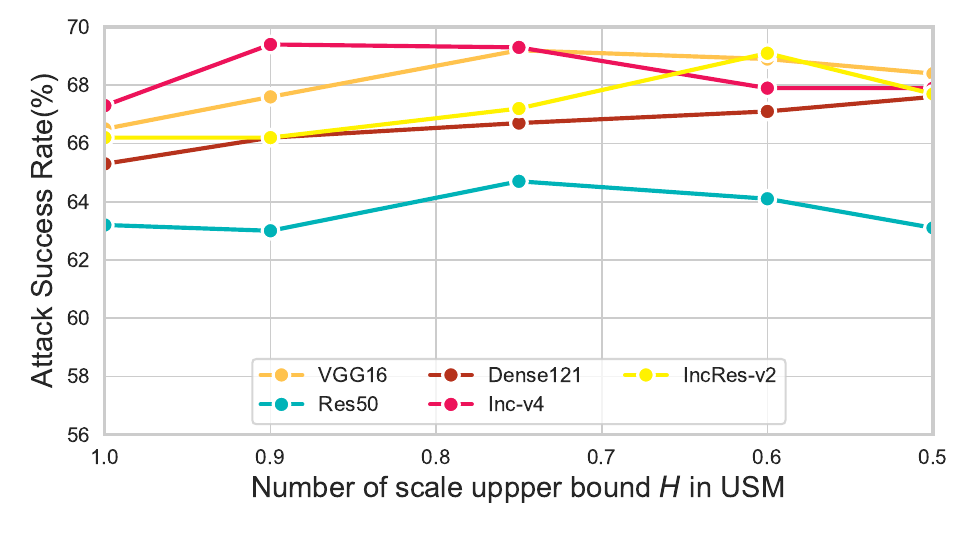}}
	\caption{Attack success rates (\%) of USM when attacking other five pretrained models for different scale upper bound $H$. }
	\label{Fg_5_upper}
\end{figure}

\subsubsection{Uniform Scale Method} \indent

\textbf{Lower bound of scale}. Considering scale copies are usually not too closed to the source image, we set upper bound to $0.75$, which is $H=0.75$. 
We test different lower bound values from $0$ to $0.3$.
Experimental results are shown in Figure \ref{Fg_5_lower}.
It can be observed from Figure \ref{Fg_5_lower} that USM achieves the obviously max attack success rate when $L$ is set to $0.1$ under the condition of $H=0.75$.

\textbf{Upper bound of scale}. 
Following the study for the lower bound, experiment exploring upper bound set the lower bound to $0.1$, which refers to $L = 0.1$.
We test different upper bound values from $0.5$ to $1$. 
Experimental results are shown in Figure \ref{Fg_5_upper}.
It can be observed from Figure \ref{Fg_5_upper} that most curves are relate even, which means $H$ might not have great influence on adversarial transferability in a certain extent.
It seems that there is not an obviously outperformed upper bound value according to Figure \ref{Fg_5_upper}.
However, $H = 1$ seems not a good choice because the attack success rate is lower compared with other values.
$H=1$ means USM calculate the gradient of raw adversarial examples and it seems to occur the overfitting problem.
This might be the reason why adversarial transferability is not good enough when  $H = 1$.
Based on these results, it seems that $H = 0.75$ is a good choice when $L = 0.05$.

\begin{figure}[tbp]
	\centerline{\includegraphics[width=1.0\columnwidth]{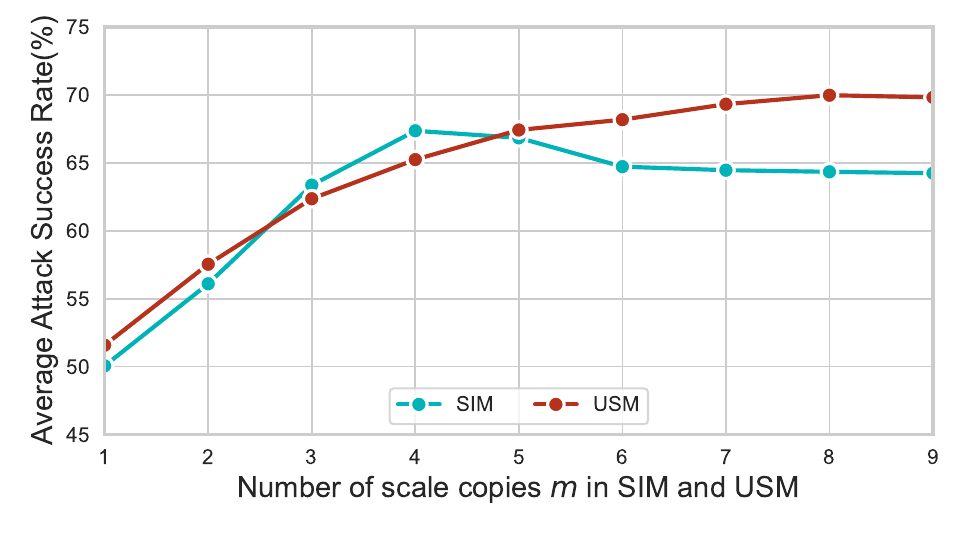}}
	\caption{Average attack success rates (\%) of USM and SIM when attacking five pretrained models.}
	\label{Fg_5_SIM_USM}
\end{figure}

\textbf{USM vs. SIM}. To validate the effect of USM, we compare USM with SIM in the setting of different $m$, which denotes the number of scale copies. We set $L=0.1$ and $H=0.75$ for USM in the comparative experiment based on the performance found in the above two ablation experiments. In Figure \ref{Fg_5_SIM_USM}, it can be observed that USM has the similar attack performance with SIM when $m$ increases from 1 to 5. However, when $m$ keeps increasing, SIM achieves a reduced attack success rate while our USM still has an increasing attack success rate.

\begin{figure}[tbp]
	\centerline{\includegraphics[width=1.0\columnwidth]{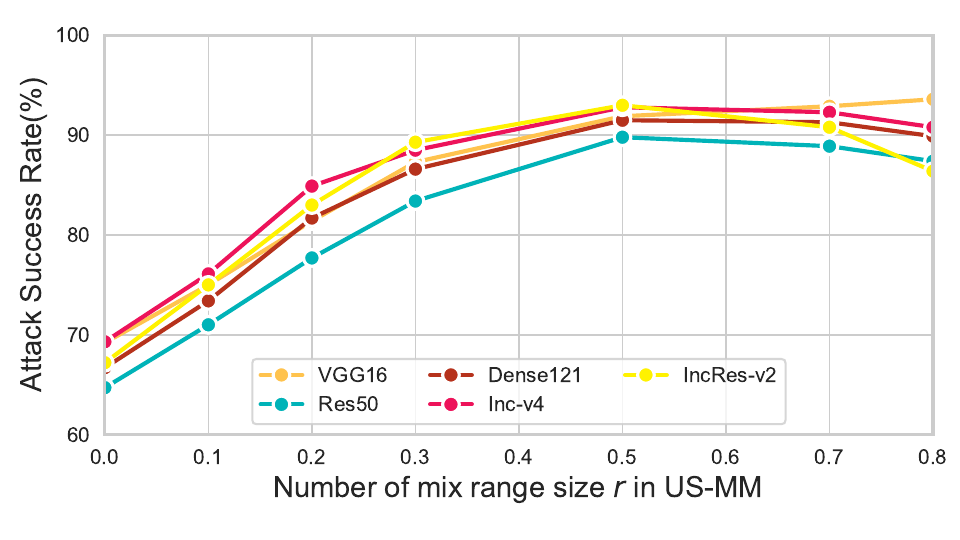}}
	\caption{Attack success rates (\%) of US-MM when attacking other five pretrained models for different mix range size $r$.}
	\label{Fg_5_range}
\end{figure}

\begin{figure}[tbp]
	\centerline{\includegraphics[width=0.95\columnwidth]{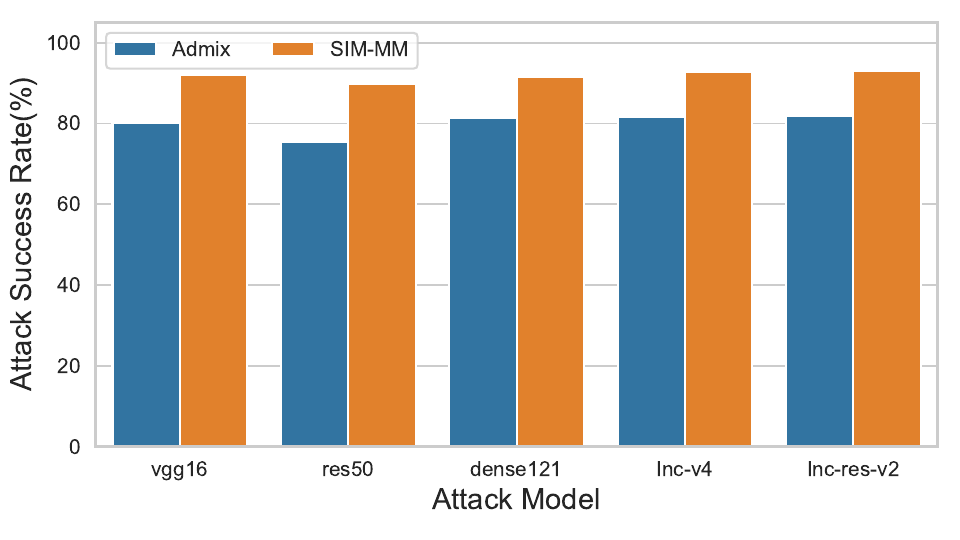}}
	\caption{Attack success rates (\%) of SIM-MM and Admix when attacking five pretrained models.}
	\label{Fg_5_SIM_MM}
\end{figure}

\subsubsection{Mix Mask Method} \indent

\textbf{Mix range size}. To investigate the relationship between attack success rate and mix range size $r$, we conduct experiments with getting $r$ from 0 to 0.8. We set $L=0.1$ and $H=0.75$ as above studies. As shown in Figure \ref{Fg_5_range}, the attack success rate increases rapidly when $r$ is set from 0 to 0.5. Then the transferability seems have a little decrease when $r$ becomes bigger than 0.5. 
It seems that a smaller $r$ value results in a smaller transformation magnitude, which leads to a crafted image with a similar gradient as the original image. On the other hand, a larger $r$ value can destroy the features of the original image and introduce harmful gradient information. This highlights the importance of finding a balance between the two for optimal adversarial transferability.

\textbf{SIM-MM vs. Admix}. For demonstrating the advantage of MM, we integrate MM to SIM as SIM-MM and conduct the comparison between SIM-MM and Admix. 
For Admix, $m_1$ and $m_2$ are set to 5 and 3 respectively. 
To maintain the same computational complexity with Admix, SIM-MM is done in the setting of $m=5$ and $m_{mix}=3$. 
Experimental results are shown in figure \ref{Fg_5_SIM_MM}. It can be observed that with the same scale strategy, SIM-MM shows a much better attack performance than Admix on five test models.

\subsection{Attack Transferability} \indent

In this section, we apply our baseline and proposed attack methods on six target models. To ensure the same computational complexity as Admix and SIT, we set the number of uniform scale copies to $m_{us}=5$ and the number of mix images to $m_{mix}=3$ in US-MM. For our ablation experiment, we set the scale lower bound $L=0.1$, scale upper bound $H=0.75$, and mix range size $r=0.5$. We then collect the results of the model outputs for the generated adversarial examples and count the number of images that are incorrectly classified. The attack success rate is defined as the proportion of these images to the entire dataset. The experimental results are shown in Table \ref{big_table}.

It can be observed that our proposed method, US-MM, achieves the best performance in almost all situations. In the two cases where it does not have the highest attack success rate, it is very close to the best. Additionally, when thoroughly examining the results, it can be seen that US-MM has a significant improvement compared to the second-best attack success rate, with an average increase of 7\%. Overall, the comparison and ablation experiment demonstrate that the combination of USM and MM in US-MM leads to a further improvement in adversarial transferability.
\section{Conclusion} \indent

In this paper, we propose a novel adversarial example generation method, namely US-MM Method. 
US Method refines the scale changing with uniforming the scale changes within the scope between the upper bound and lower bound.
MM Method improves the mixup strategy from linear addition to a mix mask method considering value range for both source image and mix image.
MM Method also considers the impact from mix image in both positive and negative direction.
Ablation experiment explores the influence of the hyper-parameter and verifies the effectiveness of both US Method and MM Method.
The results of the comparison experiment clearly demonstrate the superior performance of our proposed method in adversarial transferability.
In the future, we attempt to theoretically analyze the characteristics of adversarial transferability from three directions: model space, sample space, and feature space, and provide more detailed explanations. 
{
	\small
	\bibliographystyle{ieeenat_fullname} 
	\bibliography{main}
}


\end{document}